\documentclass[11pt]{article}
\usepackage{geometry}
\usepackage{coling2020}
\usepackage{times}
\usepackage{url}
\usepackage{latexsym}
\usepackage{microtype}
\usepackage{graphicx}
\usepackage{verbatim}
\usepackage[utf8]{inputenc}
\usepackage[T1]{fontenc}

\colingfinalcopy % Uncomment this line for all SemEval submissions

\title{Deep Learning Brasil - NLP at SemEval-2020 Task 9: Overview of Sentiment Analysis of Code-Mixed Tweets}

\author{
    Manoel Veríssimo dos Santos Neto\\
    INF - Federal University of Goiás \\
    {\tt verissimo.manoel@gmail.com} \\\And
    Ayrton Denner da Silva Amaral \\
    INF - Federal University of Goiás \\
    {\tt ayrtondenner2013@gmail.com} \\\AND
    Nádia F. F. da Silva \\
    INF - Federal University of Goiás \\
    {\tt nadia@inf.ufg.br} \\\And
    Anderson da Silva Soares \\
    INF - Federal University of Goiás \\
    {\tt anderson@inf.ufg.br} \\
}

\date{}

\begin{document}
\maketitle
\begin{abstract}
    In this paper, we describe a methodology to predict sentiment in code-mixed tweets (hindi-english). Our team called \textbf{verissimo.manoel} in CodaLab\footnote{https://competitions.codalab.org/competitions/20654} developed an approach based on an ensemble of four models (MultiFiT, BERT, ALBERT, and XLNET). The final classification algorithm was an ensemble of some predictions of all softmax values from these four models. This architecture was used and evaluated in the context of the SemEval 2020 challenge (task 9), and our system got \textbf{72.7\%} on the F1 score.
\end{abstract}

\section{Introduction}
    It is a common tendency among multilingual people who are non-native English speakers to code-mix in their speech using English-based phonetic typing. This linguistic phenomenon, particularly in social media like Twitter\footnote{https://twitter.com/}, poses a great challenge to the conventional Natural Language Processing (NLP) study area.

    Within the context of the Sentiment Analysis, the study of the phenomenon of code-mixed language is important to the research community because this behavior is more common today. The interest in this area has grown due to the volume of data that social networks generate, and also by the value that this information has to understand people opinions when they are expressed in written texts.
    
    In this paper, we explain our methodology to predict sentiment in tweets, describing how our method is based on a combination of the latest language models, and also how such models contributed to a great advance in this task. This configuration was employed and evaluated in the SemEval 2020 challenge (task 9), in which the goal is to predict the sentiment in code-mixed texts written in English and Hindi languages of a tweet \cite{patwa2020sentimix}. The models used in this combination are: MultiFiT \cite{eisenschlos2019MultiFiT} that an evolution of ULMFiT \cite{howard2018universal}, BERT \cite{devlin2018bert}, ALBERT \cite{lan2019albert} and XLNet \cite{yang2019xlnet}.

    This work is organized as follows: Section \ref{section:related_works} explains some related works, Section \ref{section:dataset_and_task} describes the dataset used, Section \ref{section:methodology} addresses the methodology applied in the task, Section \ref{section:results} presents the results, and finally Section \ref{section:conclusion} expose our final considerations as well as possible future works.

\section{Related Works}
\label{section:related_works}
    Sentiment Analysis in Twitter has been considered as a very important task from various academic and commercial perspectives. Many companies use the data on Twitter to decide about marketing and business decisions.
    
    A challenge is to apply Sentiment Analysis in texts written in two different languages like English and Hindi. This happens because some people around the world speak two or more languages and sometimes when these people write texts they write it in more than one language.
    
    The author \cite{sarkar2018jukssailcodemixed2017} present their work for Sentiment Analysis for Indian Languages (SAIL) (code mixed). They implemented an algorithm using Multinomial Naïve Bayes trained using n-gram and SentiWordnet features.
    
    For \cite{7732099}, users tend to express their thoughts by mixing words from multiple languages, because in most of the time, they are comfortable in their regionalistic language, and mixed languages are common when users write in social media. They divide their technique into two stages, viz Language Identification, and Sentiment Mining Approach. They evaluated their results and compared to a baseline obtained from machine-translated sentences in English, and found it to be around 8\% better in terms of precision.
    
    For \cite{ghosh2017sentiment} is very important a preprocessing step to remove noise from raw text. The authors developed a Multilayer Perceptron to determine sentiment polarity in code-mixed social media text from Facebook. Such example using texts from Facebook is important to show that this phenomenon can be applied in different social media than the one used on SemEval 2020.

    \cite{patra2018sentiment} cites some data about the Census in India, showing the existence of 22 scheduled languages and 462 million users on the internet. To express their feelings, such users probably use more than one language to write text on social media.
    
    In order to help NLP researchers, a corpus was created by \cite{swami2018corpus} using English-Hindi code-mixed of tweets marked for the presence of sarcasm and irony where each token is also annotated with a language tag. In this present work, this corpus was used to training MultiFiT model.

\section{Methodology}
\label{section:methodology}

\begin{figure}[!ht]
\centering
\includegraphics[width=0.70\textwidth]{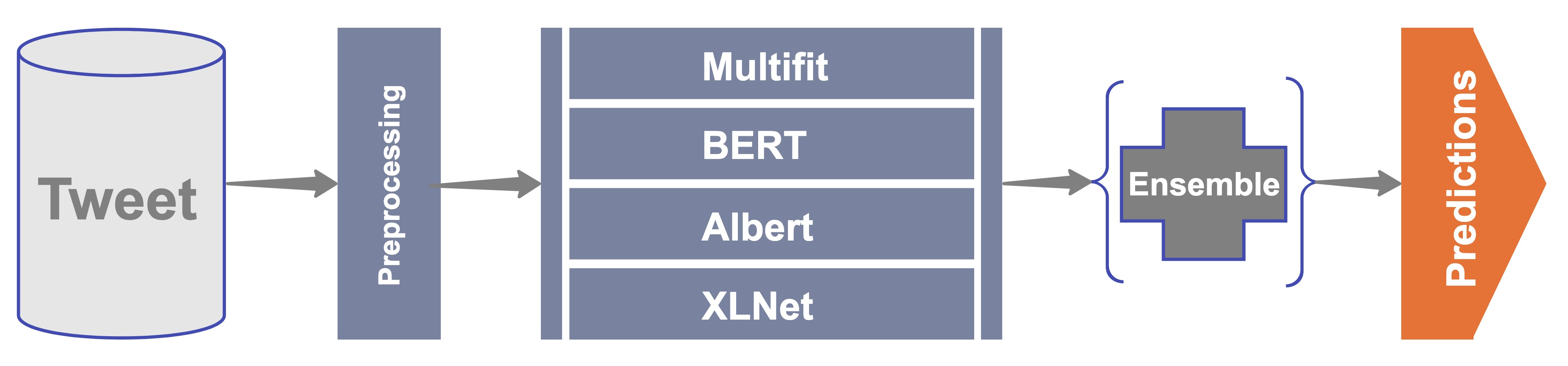}
\caption{Solution Architecture.}
\label{fig:architecture}
\end{figure}

    The methodology applied in this task consists of training and using prediction values of four models: MultiFiT, BERT, ALBERT, and XLNet. After retrieving prediction values, our ensemble calculates an average of all softmax values from these four models, as shown in Figure \ref{fig:architecture}. Models BERT, ALBERT, and XLNet were trained on a DGX-1, while MultiFiT was trained on a GTX 1070 Ti 8GB. The hyperparameters of the four models are described in Table \ref{Table_Hyperparameters}.
    
    \begin{table}[!ht]
        \small
        \centering
        \begin{tabular}{lllllll}
            \hline
            \bf Model & \bf Batch Size & \bf Learning Rate & \bf Max Length & \bf Optimizer & \bf Training & \bf Base Model \\ \hline
            \\
            MultiFiT & 32 & 1e-3 &  &  & 20 epochs &  \\

            BERT & 8 & 2e-5 & 128 & AdamW & 30 epochs & BERT-Base, Multilingual Cased \\

            ALBERT & 16 & 2e-5 & 256 & AdamW & 30 epochs & Xxlarge \\

            XLNet & 16 & 2e-5 & 256 &  & 8000 steps & XLNet\-Large, Cased \\

            \\
            \hline
        \end{tabular}
        \caption{Hyperparameters}
        \label{Table_Hyperparameters}
    \end{table}

\subsection{Preprocessing}

    This step consists in eliminating noises and terms that have no semantic significance in the sentiment prediction. For this, we perform the removal of links, removal of numbers, removal of special characters, and transform text in lowercase.

\subsection{MultiFiT}

    Nowadays, there are many advances in NLP, but the majority of researches is based on the English language, and those advances can be slow to transfer beyond English.
    
    The \textbf{MultiFiT} \cite{eisenschlos2019MultiFiT} method is based on \textbf{Universal Language Model Fine-tuning (ULMFiT)} \cite{howard2018universal} and the goal of this model is to make it more efficient for modeling languages others than English.
    
\begin{figure}[!ht]
\centering
\includegraphics[width=0.80\textwidth]{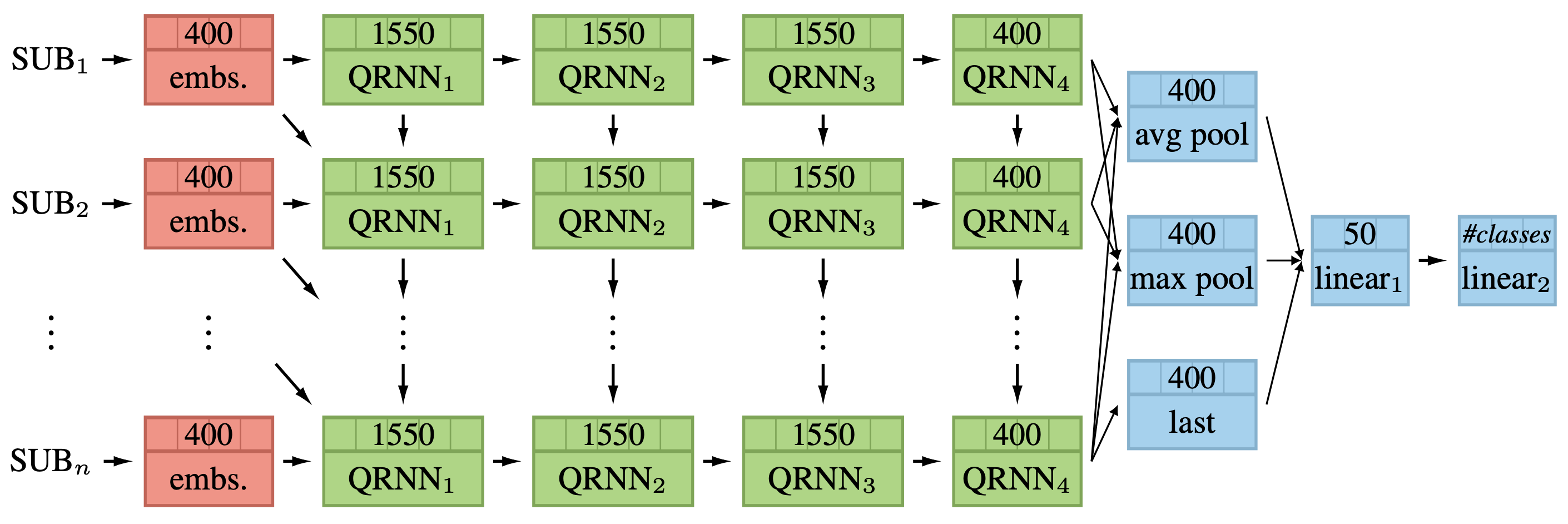}
\caption{MultiFiT Architecture.\footnotemark}
\label{fig:MultiFiT_architecture}
\end{figure}

    There are two changes compared to the old model: it utilizes tokenization based on sub-words rather than words, and it also uses a QRNN \cite{bradbury2016quasirecurrent} rather than an LSTM. The model architecture can be seen in Figure \ref{fig:MultiFiT_architecture}.
    
    The architecture of the model consists of a subword embedding layer, four QRNN layers, an aggregation layer, and two linear layers. In special this architecture, subword tokenization has two very important properties:
    
    \begin{itemize}
        \item Subwords more easily represent inflections and this includes common prefixes and suffixes. For morphologically rich languages this is well-suited.
        \item It is a common problem out-of-vocabulary tokens and Subword tokenization is a good solution to prevent this problem.
    \end{itemize}

\footnotetext{\url{https://nlp.fast.ai/classification/2019/09/10/multifit.html}}

\subsection{BERT}

    \textbf{B}idirectional \textbf{E}ncoder \textbf{R}epresentations from \textbf{T}ransformers (also abbrevitaed as \textbf{BERT}) \cite{devlin2018bert} is a model designed to pre-train deep bidirectional representations from unlabeled data. The pre-trained BERT model can be fine-tuned with just one single additional output layer, which can be used in sentiment analysis and others NLP tasks.
    
    The implementation of BERT there has two steps: pre-training and fine-tuning. In the pre-training step, the model is trained on unlabeled data over different pre-training tasks using a corpus in a specific language or in multiples corpus with different languages. For the fine-tuning step, the BERT model is first initialized with the pre-trained parameters, and all of the parameters are fine-tuned using labeled data from the specific tasks.
    
    Dataset repositories like NLP-progress\footnotemark\space track different model results and progress in many Natural Language Processing (NLP) benchmarks, and also the current state for the most common NLP tasks. When doing a comparison between results available for reference in such repositories, BERT was able to achieve state-of-the-art in many NLP-related tasks, which gives an excellent reason to use BERT in our architecture, even while many reasons of BERT state-of-art performance are not fully understood \cite{kovaleva-etal-2019-revealing} \cite{clark-etal-2019-bert}.

\footnotetext{\url{http://nlpprogress.com/english/sentiment_analysis.html}}

\subsection{ALBERT}

    Recent language models had shown a tendency to increase in size and quantity of parameters for training. They often offer many improvements in many NLP tasks, but they suffer as a consequence of the need for many hours of training, which consequently increases its costs of operation. \textbf{ALBERT} \cite{lan2019albert}: \textbf{A} \textbf{L}ite \textbf{BERT} for Self-supervised Learning of Language Representations, offers an alternative of parameters reduction to solve this problem.
    
    There are two changes to reduce the size of the model based on BERT. The first is a factorized embedding parametrization, this decomposing the large vocabulary embedding matrix into two small matrices.This decomposition approach reduces the trainable parameters and reduces a significant time during the training phase. The second change is the share parameter cross-layer, which also prevents the parameter from growing with the depth of the network.
    
\subsection{XLNet}

    \textbf{XLNet} \cite{yang1906xlnet} is a model that uses a bidirectional learning mechanism, doing that as an alternative to word corruption via masks implemented by \textbf{BERT}. XLNet uses a permutation operation over tokens in the same input sequence, being able to use a single phrase through different training steps while providing different examples. Phrase permutation in training fixes token position, but iterating in every token in training phrases, rendering the model able to deal with information gathered from tokens and its positions in a given phrase. XLNet also draws inspiration from \textbf{Transformer-XL} \cite{dai2019transformer}, relying specially in pretraining ideas.

\section{Dataset and Task}
\label{section:dataset_and_task}
\begin{figure}[!ht]
\centering
\includegraphics[width=0.80\textwidth]{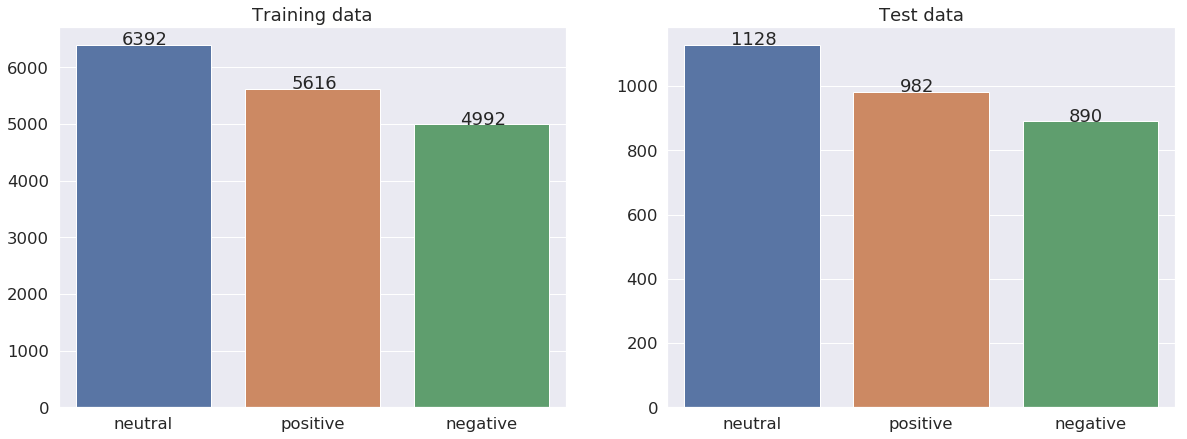}
\caption{Number of labels per classes.}
\label{fig:classes}
\end{figure}

\noindent
\textbf{Dataset.} The data for the task consists of 17,000 tweets for training and 3,000 for test/evaluation. The data format is in CONLL format. The amount of tweets for the dataset can be seen in Figure \ref{fig:classes}.

\vspace{5mm}

\noindent
\textbf{Task details.} The task objective is to predict the sentiment of a given code-mixed tweet. The sentiment labels are positive, negative, or neutral, and the code-mixed languages will be English-Hindi. The challenge is to predict the sentiment in texts written in these two languages \cite{patwa2020sentimix}.

Here are some examples of sentences taken from the dataset.
\vspace{5mm}

\textbf{Positive Sentence}
\begin{quote}
    \textit{@AmitShah @narendramodi All India me nrc lagu kare w Kashmir se dhara 370ko khatam kare ham Indian ko apse yahi umid hai}
\end{quote}

\textbf{Negative Sentence}
\begin{quote}
    \textit{@RahulGandhi television media congress ke liye nhi h . Ye toh aapko pata chal hi gya hoga . Achha hoga ki Congress ke … https//t.co/HmH8M7PTaK}
\end{quote}

\textbf{Neutral Sentence}
\begin{quote}
    \textit{@sardanarohit jaaz saab ko salo saal ke pending case ko soultion me maza nahi aata * inko to public paise monthly case miljaye *}
\end{quote}

\section{Results}
\label{section:results}
    In this section, we report the obtained results by our model according to the metric evaluation used by the challenge: macro f1, precision and recall, accuracy, and f1 for all classes. Results are reported for each model and an ensemble using a combination of results of the four models XLNet, BERT, ALBERT, and MultiFiT. In Table \ref{Table1} we show model’s performances and in Table \ref{Table2} we present the F1 score per class.
    
    \begin{table}[!ht]
        \centering
        \begin{tabular}{lllll}
            \hline
            \bf Model & \bf F1 & \bf P & \bf R & \bf Acc \\ \hline
            \\
            \bf Ensemble & \bf 0.727 & \bf 0.729 & \bf 0.726 & \bf 0.723 \\ 
            XLNet & 0.679 & 0.696 & 0.692 & 0.690 \\ 
            ALBERT & 0.679 & 0.684 & 0.676 & 0.675 \\
            BERT & 0.675 & 0.680 & 0.672 & 0.670 \\
            MultiFiT & 0.665 & 0.665 & 0.669 & 0.662 \\
            \\
            \hline
        \end{tabular}
        \caption{Result Semeval-2020.}
        \label{Table1}
    \end{table}
    
    \begin{table}[!ht]
        \centering
        \begin{tabular}{ |c|c|c|c| } 
            \hline
            \bf & \bf Class & \bf Acc \\
            \hline
            \includegraphics[scale=0.12]{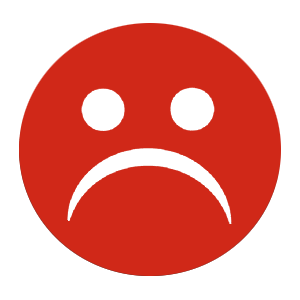} & Negative & 0.671 \\ 
            \includegraphics[scale=0.12]{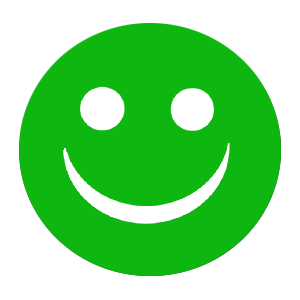} & Positive & \bf 0.760 \\ 
            \includegraphics[scale=0.12]{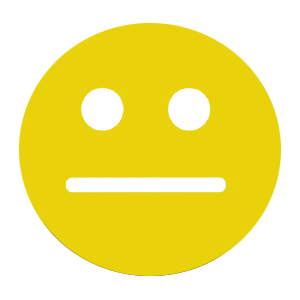} & Neutral & 0.606 \\ 
            \hline
        \end{tabular}
        \caption{Ensemble f1 by class.}
        \label{Table2}
    \end{table}
    
    The obtained results on the testing data indicate that our ensemble produces the best F1; on the other hand, the XLNet model represents the best result among the other models. It is important to quote that the final ensemble is a large combination of results of the four models used in this architecture.
    
    For the official\footnote{https://competitions.codalab.org/competitions/20654\#learn\_the\_details-results} results in competition, the organizers used only the first three submissions and in our case, our models were only MultiFiT and BERT. Using only these architectures our results are only \textbf{66.5\%}.

\section{Conclusion}
\label{section:conclusion}
    In this paper, we propose a combination of four models for the Semeval 2020 (task 9), and our team got \textbf{72.7\%} on F1 score in the competition. All of these models are based on using language models and transfer learning. They alone performed well, but together in an ensemble combination, they performed even better.
    
    In some applications, it is difficult to use an ensemble consisted of four models, especially because of the overhead coming from time spent on inference, culminating in an approach that sometimes will not perform well. On the other hand, the individual results of these four models are very close, meaning that for this task, any model can be used.
    
    It is important to note that MultFit has the worst result, but the difference is very small, and this specific model takes a lot less time to train, being the lightest model of the ensemble.
    
    As future works, we intend to explore these models for Sentiment Analysis in other multilingual and monolingual scenarios.

\newpage

\bibliographystyle{coling}
\bibliography{semeval2020}

\end{document}